\documentclass[10pt]{article}

\usepackage[utf8]{inputenc}

\usepackage[margin=1.1in]{geometry}

\usepackage{natbib}
\usepackage[colorlinks=true,citecolor=blue]{hyperref}%

\RequirePackage[OT1]{fontenc}

\usepackage{amssymb,amsmath,amsthm}
\usepackage{bbold}

\usepackage{algorithm}
\usepackage{algpseudocode}

\usepackage{graphicx}

\usepackage{tgheros} 

\newcommand{\rf}{\mathrm{ref}}
\newcommand{\old}{\mathrm{old}}

\usepackage{mathpazo}

\title{{\bf What is the Alignment Objective of GRPO?}}

\author{Milan Vojnovic and Se-Young Yun\thanks{M. Vojnovic is with the Department of Statistics, London School of Economics, London, UK, \url{m.vojnovic@lse.ac.uk}. S. Yun is with KAIST AI, Seoul, South Korea, \url{yunseyoung@kaist.ac.kr}.}}

\date{}

\begin{document}

\maketitle

\begin{abstract} In this note, we examine the aggregation of preferences achieved by the Group Relative Policy Optimisation (GRPO) algorithm, a reinforcement learning method used to train advanced artificial intelligence models such as DeepSeek-R1-Zero and DeepSeekMath \citep{deepseekai2025,shao2024deepseekmathpushinglimitsmathematical}. The GRPO algorithm trains a policy using a reward preference model, which is computed by sampling a set of outputs for a given context, observing the corresponding rewards, and applying shift-and-scale normalisation to these reward values. Additionally, it incorporates a penalty function to discourage deviations from a reference policy.

We present a framework that enables us to characterise the stationary policies of the GRPO algorithm. This analysis reveals that the aggregation of preferences differs fundamentally from standard logarithmic pooling, which is implemented by other approaches such as RLHF. The precise form of preference aggregation arises from the way the reward preference model is defined and from the penalty function, which we show to essentially correspond to the reverse Kullback–Leibler (KL) divergence between the aggregation policy and the reference policy.

Interestingly, we demonstrate that for groups of size two, the reward preference model corresponds to pairwise comparison preferences, similar to those in other alignment methods based on pairwise comparison feedback. We provide explicit characterisations of the aggregate preference for binary questions, for groups of size two, and in the limit of large group size. This provides insights into the dependence of the aggregate preference on parameters such as the regularisation constant and the confidence margin of question answers.

Finally, we discuss the aggregation of preferences obtained by modifying the GRPO algorithm to use direct KL divergence as the penalty or to use rewards without scale normalisation.    
\end{abstract}

\section{Introduction}
\label{sec:intro}

The recently developed advanced artificial intelligence model, DeepSeek R1, has demonstrated remarkable performance in solving complex reasoning tasks, logic problems, and other step-by-step problems \cite{deepseekai2025}. At its core, the system employs reinforcement learning, specifically the Group Relative Policy Optimisation (GRPO) algorithm, originally proposed in \cite{shao2024deepseekmathpushinglimitsmathematical}. The objective is to train a language model using reinforcement learning, where feedback preferences serve as a reward signal alongside a reference language model. This process can be viewed as aligning the reward maximisation preference and the reference model's preference. For example, the rewards may be accuracy indicators computed through a rule-based reward system, determining whether a response to a given question is correct. In this note, we refer to responses as outputs and questions as contexts. GRPO extends the previously proposed Proximal Policy optimisation (PPO) algorithm \citep{schulman2017ppo} in several ways: it introduces a novel method for computing the advantage of outputs in a given context by sampling a group of outputs, and it incorporates a new regulariser based on an estimator of the Kullback-Leibler (KL) divergence to bias the policy towards a reference policy model. 

In this note, we examine the GRPO algorithm, focusing on its alignment properties and their relationship to other alignment algorithms. 

{\bf Group Relative Policy optimisation (GRPO)} For a context $q$ sampled from a distribution $\mu$, the GRPO algorithm samples a group of outputs $o_1,\ldots, o_G$ from an old policy $\pi_{\theta_{\old}}(\cdot \mid q)$, observes their corresponding rewards $r_1, \ldots,r_G$, and uses this information, along with a given reference policy $\pi_\rf(\cdot\mid q)$, to define the objective function for selecting a new policy. This new policy $\pi_\theta$ aims to maximise the following objective:
\begin{equation}
\mathcal{J}_{GRPO}(\theta)=\mathbb{E}_{q\sim \mu, \{o_i\}_{i=1}^G \stackrel{\text{i.i.d.}}{\sim} \pi_{\theta_{\old}}(\cdot\mid q)}\left[\frac{1}{G}\sum_{i=1}^G \left(\tilde{A}_i(\theta)  -\beta D_i(\theta)\right)\right]
\label{equ:Jgrpo}
\end{equation}
with
$$
\tilde{A}_i(\theta) = \min\left\{\frac{\pi_\theta(o_i\mid q)}{\pi_{\theta_{\old}}(o_i\mid q)}A_i, \mathrm{clip}_\epsilon\left(\frac{\pi_\theta(o_i\mid q)}{\pi_{\theta_{\old}}(o_i\mid q)}\right)A_i\right\},
$$
$$
D_i(\theta) = \frac{\pi_{\rf}(o_i\mid q)}{\pi_\theta(o_i\mid q)} - \log \frac{\pi_{\rf}(o_i\mid q)}{\pi_\theta(o_i\mid q)} -1,
$$
where $G$ is an integer-valued hyperparameter greater than or equal to two, and $\epsilon$ and $\beta$ are positive-valued hyperparameters. 
The function $\mathrm{clip}_\epsilon(x)$ outputs $x$ if $1-\epsilon\leq x \leq 1+\epsilon$, $1-\epsilon$ if $x<1-\epsilon$, and $1+\epsilon$ if $x>1+\epsilon$.
Additionally, $A_i$ represents the advantage corresponding to the output $o_i$ within the group, defined as:
\begin{equation}
A_i = \frac{r_i - \mathrm{mean}(r_1,\ldots, r_G)}{\mathrm{std}(r_1,\ldots,r_G)}.
\label{equ:ai}
\end{equation}
Here, $\mathrm{mean}(r_1,\ldots,r_G)\stackrel{\mathrm{def}}{=}(1/G)\sum_{i=1}^G r_i$ and $\mathrm{std}(r_1,\ldots, r_G) \stackrel{\mathrm{def}}{=} \sqrt{(1/G)\sum_{i=1}^G (r_i - \mathrm{mean}(r_1,\ldots,r_G))^2}$. In Equation~(\ref{equ:ai}), we define $A_i = 0/0 \equiv 0$ in the case where $r_1 = \cdots = r_G$.

The objective in (\ref{equ:Jgrpo}) consists of two terms: a \emph{reward preference model} and a \emph{reference-policy divergence penalty}. The reward preference model is designed to favour outputs that achieve a higher reward relative to other outputs within the group. The reference-policy divergence penalty discourages policies from deviating excessively from the reference policy. The decomposition of the objective function into a reward preference model and a reference-policy divergence penalty is a common approach in various alignment algorithms, with key differences arising in how these terms are specifically defined.

The definition of the advantage values in Equation~(\ref{equ:ai}), which is equivalent to
$$
A_i = \frac{\frac{1}{G}\sum_{j=1}^G (r_i - r_j)}{\sqrt{\frac{1}{2G^2}\sum_{j=1}^G \sum_{k=1}^G (r_j - r_k)^2}},
$$
applies shift and scale normalisation. Using shift normalisation with a baseline is a standard technique for variance reduction in reinforcement learning, particularly when the average reward is used as the baseline \citep{suttonbarto}. While scale normalisation is perhaps less common in reinforcement learning, both shift and scale normalisation are widely used in machine learning. Importantly, these normalisation techniques ensure that the advantage terms remain invariant under shift and scale transformations of the input rewards. 

Both the reward preference model and the reference-policy divergence penalty of GRPO differ from those used in some well-known alignment approaches, which we discuss in the following section.

\subsection{Related work}


In this section, we review two existing alignment approaches, which we use as baselines to compare their alignment objectives with that of the GRPO algorithm. 

\paragraph{Reinforcement Learning from Human Feedback (RLHF)} The standard RLHF paradigm \citep{Christiano2017,Stiennon2020} consists of two main steps: learning the reward model and optimising the policy using the learned reward model. In the first step, the reward model $r_\phi(\cdot\mid q)$ is trained on a dataset containing examples of human preferences in the form of pairwise comparisons of outputs for given contexts. In the second step, the objective is to find a policy that maximises the following objective function:
\begin{equation}
\mathcal{J}_{RLHF}(\theta) = \mathbb{E}_{q\sim \mu, o\sim \pi_\theta(\cdot\mid q)}[r_\phi(o\mid q)] - \beta\ \mathbb{E}_{q\sim \mu}[\mathrm{KL}(\pi_\theta(\cdot\mid q)\mid\mid \pi_{\rf}(\cdot \mid q))]
\label{equ:RLHF}
\end{equation}
where $\pi_\rf(\cdot\mid q)$ is a reference model policy and $\mathrm{KL}(\pi\mid\mid \pi')$ is the Kullback-Leibler divergence between two distributions $\pi$ and $\pi'$, $\mathrm{KL}(\pi\mid \mid \pi')=\mathbb{E}_{x\sim \pi}[\log(\pi(x)/\pi'(x))]$. The objective function in (\ref{equ:RLHF}) is optimised by using PPO \citep{schulman2017ppo} or similar optimisation approaches. PPO is an actor-critic RL algorithm that is used in the RL fine-tuning stage of LLMs \citep{Ouyang2011}.

The RLHF can be seen as an approach for aggregating a reward preference and a reference-policy preference according to:
\begin{equation}
\pi_\theta(o\mid q ) = \frac{1}{Z_q}\pi_{\rf}(o\mid q) e^{\frac{1}{\beta}r_\phi(o\mid q)}
\label{equ:rlhfpi}
\end{equation}
where $Z_q$ is a normalisation constant. This follows directly by choosing $\pi_\theta(\cdot\mid q)$ that maximises the RLHF objective function (\ref{equ:RLHF}). The aggregate preference distribution in (\ref{equ:rlhfpi}) follows the logarithmic opinion pooling form \citep{logarithmic}. Logarithmic opinion pooling is a method for aggregating multiple probability distributions into a single consensus distribution, where the consensus distribution is proportional to some weighted geometric average of the individual distributions. Specifically, (\ref{equ:rlhfpi}) defines the consensus distribution as a weighted geometric average of the reference-policy distribution $\pi_\rf(\cdot\mid q)$ and the Luce's choice \citep{luce1959individual} distribution $e^{r_\phi(o\mid q)}/\sum_{o'}e^{r_\phi(o'\mid q)}$ parametrised with the reward values, with the respective weights of values $1$ and $1/\beta$.  

\paragraph{Nash Learning from Human Feedback (NLHF)} NLHF is an alignment approach introduced in \citep{pmlr-v235-munos24a}, where the reward preference model is defined in terms of pairwise preferences over outputs for a given context. Specifically, the preference of an output $o$ over another output $o'$, given a context $q$, is expressed as a value $\mathcal{P}(o\succ o'\mid q)$ in the range $[0,1]$. The pairwise preferences are assumed to be antisymmetric, meaning that $\mathcal{P}(o'\succ o\mid q)=1-\mathcal{P}(o\succ o'\mid q)$. Given the pairwise preferences $\mathcal{P}(o\succ o'\mid q)$ and a reference policy $\pi_\rf(\cdot\mid q)$ for each context $q$, the aggregation of preferences is defined as the symmetric Nash equilibrium of a two-player zero-sum game. The expected payoff for a player deploying the mixed strategy $\pi$ against a player deploying the mixed-strategy $\pi'$, is given by:
\begin{equation}
\mathcal{J}_{NLHF}(\pi,\pi') = \mathbb{E}_{q\sim \mu, o\sim \pi(\cdot\mid q), o'\sim \pi'(\cdot\mid q)}\left[\mathcal{P}(o\succ o'\mid q) - \beta \log\frac{\pi(o\mid q)}{\pi_\rf(o\mid q)} + \beta \log\frac{\pi'(o'\mid q)}{\pi_\rf(o'\mid q)}\right]
\label{equ:nlhf}
\end{equation}
where $\beta$ is a positive-valued hyperparameter. The NLHF two-player zero-sum game has a unique Nash equilibrium, which is also the limit point of a mirror-descent iterative computation algorithm \citep{pmlr-v235-munos24a}. 

A notable difference between NLHF and RLHF is that NLHF observes reward preferences as pairwise comparisons of outputs, whereas RLHF expresses preferences as absolute reward vaues assigned to individual outputs. It can be easily shown that the solution to the NLHF game satisfies:
$$
\pi(o\mid q) = \frac{1}{Z_q}\pi_\rf(o\mid q) e^{\frac{1}{\beta}\mathbb{E}_{o'\sim \pi(\cdot\mid q)}[\mathcal{P}(o\succ o'\mid q)]}
$$
where $Z_q$ is a normalisation constant. Notably, this can also be interpreted as a logarithmic pooling of distributions, where the geometric averaging weights depend on $\pi$.

\subsection{Summary of our findings}

Our findings can be summarised in the following points:

\begin{itemize}
\item We present a framework for analysing the stationary policies of the GRPO algorithm, expressing the reward preference model and the reference-policy divergence penalty in a way that reveals their fundamental role in aligning preferences. This framework clarifies the contribution of individual components and their relationship to previously proposed algorithms for preference aggregation.  

\item We show that preference aggregation in GRPO corresponds to scaling the reference probability of an output, given a context, by a function that increases with the expected advantage of the output relative to the expected advantage of a randomly chosen group of outputs from the aggregate probability distribution. This form of preference aggregation differs from the logarithmic pooling used in methods such as RLHF. 

\item For groups of size two, we show that the reward preference model corresponds to pairwise comparison preferences, where comparisons involve point rewards for outputs in a pair, given a context. In the limit of large group sizes, the reward preference model converges to the expected reward normalised by the standard deviation of the reward of an output sampled from the previous policy.

\item Regarding the reference-policy divergence penalty, we find that for stationary policies, GRPO’s penalty is essentially equivalent to the reverse KL divergence between the new candidate policy and the reference policy. It is unclear whether this was the intended design, as the penalty was originally motivated as an estimator of the direct KL divergence. The fact that the penalty effectively corresponds to the reverse KL divergence plays a key role in shaping how preferences are aggregated. 

\item We derive explicit closed-form expressions for the stationary policies in the case of binary questions, for groups of size two, and in the asymptotic limit of large groups. Preference aggregation follows a nonlinear transformation of the reference probability distribution, favouring the more rewarding answer. Notably, for groups of size two, the more rewarding answer is guaranteed to have an aggregate probability at least as large as a value dependent solely on the ratio of the regularisation constant to the confidence margin of the question answers---approaching 1 for small values of this ratio. In the limit of large groups, this dependence reduces to the regularisation constant alone. This suggests that for practical choices of the regularisation constant, such as the default value of 0.04 used in TRL (Transformer Reinforcement Learning) by Hugging Face\footnote{\url{https://huggingface.co/docs/trl/main/en/grpo_trainer\#trl.GRPOConfig}}, preference aggregation may predominantly reflect the reward preference.

\item Finally, we discuss the implications of adjusting the reference-policy divergence penalty to align with the direct KL divergence or using only shift normalisation for the reward preference model. The former adjustment results in logarithmic opinion pooling, and we present an example where the aggregate preference may not be unique. The latter adjustment aligns the reward preference with that of RLHF. Combining both adjustments leads to an aggregation of preferences consistent with the principles of RLHF.
\end{itemize}

\subsection{Additional assumptions}

For the optimisation problem to be well defined, we make the following assumptions. For every context $q$, the domain of the distribution $\pi_\theta(\cdot \mid q)$ is assumed to be contained within the support of the distribution $\pi_\rf(\cdot\mid q)$. This ensures that for every output $o$ in the domain of $\pi_\theta(\cdot\mid q)$, we have $\pi_\rf(o\mid q) > 0$. Without this condition, the reference-policy divergence penalty would become infinite whenever $\pi_\rf(o\mid q) = 0$ and $\pi_\theta(o \mid q) > 0$ for some output $o$ and context $q$. 

Our study focuses on characterising the stationary policies of the GRPO algorithm; therefore, we ignore the clipping function in the GRPO objective function. A stationary policy is a collection of distributions $\pi_{\theta^\star}(\cdot\mid q)$ such that $\theta^\star$ maximises $\mathcal{J}_{GRPO}(\theta)$ over $\theta$, assuming that $\pi_{\theta_{\old}}\equiv \pi_{\theta^*}$. Ignoring the clipping function is also justified when running the policy gradient algorithm with the GRPO objective function using a sufficiently small step size, ensuring that the new and old policies remain within an $\epsilon$-relative difference, i.e., $|\pi_\theta(o\mid q)-\pi_{\theta_{\old}}(o\mid q)|\leq \epsilon \pi_{\theta_{\old}}(o\mid q)$, for all $o$ and $q$.

\section{The alignment objective of the GRPO algorithm}

In this section, we analyse the alignment objective of the GRPO algorithm. We begin by examining the reward preference model and the reference-policy divergence penalty separately, before discussing the alignment objective as a whole. 

\subsection{The reward preference model}

We consider a more general setting than in Section~\ref{sec:intro}, where the reward $r$ is allowed to be stochastic for any given output $o$ and context $q$. Let $r(o\mid q)$ denote the expected value of the reward for output $o$ under context $q$. The case of deterministic rewards is a special case, where, for each output $o$ under a context $q$, the reward takes a deterministic value $r(o\mid q)$. Recall that we ignore the clipping function term in the objective function, as our focus is on characterising stationary policies. Hence, we consider the reward preference given by:
$$
\mathcal{R}_G(\theta\mid q)  \stackrel{\mathrm{def}}{=}  \mathbb{E}_{\{o_i\}_{i=1}^G \stackrel{\text{i.i.d.}}{\sim} \pi_{\theta_{\old}}(\cdot\mid q)}\left[\frac{1}{G}\sum_{i=1}^G \frac{\pi_\theta(o_i\mid q)}{\pi_{\theta_{\old}}(o_i\mid q)}A_i\right].
$$

The GRPO's reward preference model can be expressed as follows. Let $\mathcal{P}_G(o \mid \{o_i'\}_{i=1}^{G-1},q)$ denote the group-relative preference of output $o$ over outputs $o'_1, \ldots, o'_{G-1}$ for a given context $q$. For any conditional distribution $\pi'(\cdot\mid q)$ for a given context $q$, let $\mathcal{P}_G(o\mid \pi'(\cdot\mid q),q)$ be the expected group-relative preference of output $o$ for a given context $q$, i.e.,
$$
\mathcal{P}_G(o\mid \pi'(\cdot\mid q),q) \stackrel{\mathrm{def}}{=} \mathbb{E}_{o'_1,\ldots, o'_{G-1}\stackrel{\mathrm{i.i.d.}}{\sim} \pi'(\cdot\mid q)}[\mathcal{P}_G(o \mid \{o_i'\}_{i=1}^{G-1},q)].
$$

It can be readily observed that the GRPO's reward preference model can be expressed as:
\begin{equation}
\mathcal{R}_G(\theta\mid q)= \mathbb{E}_{o\sim \pi_\theta(\cdot\mid q)}[\mathcal{P}_G(o\mid \pi_{\theta_{\old}}(\cdot\mid q),q)]
\label{equ:reward}
\end{equation}
where, specifically,
$$
\mathcal{P}_G(o\mid \{o_i'\}_{i=1}^{G-1},q)\stackrel{\mathrm{def}}{=} \mathbb{E}\left[\frac{r_1 - \mathrm{mean}(r_1,r_2,\ldots, r_G)}{\mathrm{std}(r_1,r_2,\ldots, r_G)}\mid o_1 = o, o_2 = o_{1}', \ldots, o_G = o'_{G-1},q\right]
$$
where the expectation is with respect to the distributions of rewards given their corresponding outputs and the context. For the case of deterministic rewards, we have
$$
\mathcal{P}_G(o\mid \{o_i'\}_{i=1}^{G-1},q)= \frac{r(o\mid q) - \mathrm{mean}(r(o\mid q),r(o'_1\mid q),\ldots, r(o'_{G-1}\mid q))}{\mathrm{std}(r(o\mid q),r(o'_1\mid q),\ldots, r(o'_{G-1}\mid q))}.
$$

It is insightful to consider two extreme cases, one in which the group size is the smallest possible value of two outputs, and the other where the group size becomes asymptotically large. 

\paragraph{Groups of size two} For the case where each group consists of a pair of outputs, it can be readily verified that for every pair of outputs $o_i$ and $o_j$, the advantage terms take the following values:
$$
A_i = \mathrm{sign}(r_i-r_j) \hbox{ and } A_j = -A_i.
$$

Notably, the reward preference model accounts only for the relative preference between pairs of outputs---that is, which output in a pair has a higher reward---while remaining invariant to the absolute values of the rewards. This is due to the way the advantage terms are defined, and, in particular, normalisation by the standard deviation.   

The group-relative preference $\mathcal{P}_2(o\mid \{o'\},q)$ corresponds to the pairwise preference $\mathcal{P}(o\succ o'\mid q)$, defined as  $\mathcal{P}(o\succ o'\mid q) = \mathbb{P}[r_i > r_j\mid o_i = o, o_j = o', q]$. In the case of deterministic rewards, $\mathcal{P}(o\succ o'\mid q)$ takes the value of $1$ if $r(o\mid q) > r(o'\mid q)$ and the value of $0$ otherwise.

The general expression for the expected reward preference model, given in Equation~(\ref{equ:reward}), specialised for groups of size two, can be written as:
$$
\mathcal{R}_2(\theta) = \mathbb{E}_{q\sim \mu, o\sim \pi_\theta(\cdot\mid q), o'\sim \pi_{\theta_{\old}}(\cdot\mid q)}[\mathcal{P}(o\succ o'\mid q) - \mathcal{P}(o'\succ o\mid q)].
$$
If the pairwise preferences are asymmetric, meaning that $\mathcal{P}(o\succ o'\mid q) + \mathcal{P}(o'\succ o\mid q) = 1$, then 
$$
\mathcal{R}_2(\theta)=2\mathbb{E}_{q\sim \mu, o\sim \pi_\theta(\cdot\mid q), o'\sim \pi_{\theta_{\old}}(\cdot\mid q)}[\mathcal{P}(o\succ o'\mid q)]-1.
$$

Perhaps interestingly, we observe that the reward preference model corresponds to that of the NLHF model, as given in Equation~(\ref{equ:nlhf}), up to non-essential multiplicative and additive constants. 

\paragraph{The limit of large group size} By the law of large numbers, for $r_1,\ldots, r_G\stackrel{\mathrm{i.i.d.}}{\sim} \pi_{\theta_{\old}}(\cdot\mid q)$, we have 
$$
\lim_{G\rightarrow \infty} \mathrm{mean}(r_1, \ldots, r_G) = \mathrm{E}_{o\sim \pi_{\theta_{\old}}(\cdot\mid q)}[r(o\mid q)],
$$
and
$$
\lim_{G\rightarrow \infty} \mathrm{std}(r_1, \ldots, r_G) = \sigma(\pi_{\theta_{\old}}(\cdot\mid q)),
$$
where $\sigma(\pi_{\theta_{\old}}(\cdot\mid q))^2$ is the variance of the reward for an output according to the distribution $\pi_{\theta_{\old}}(\cdot\mid q)$. 

In the case of the limit of large group size, the reward preference model corresponds to:
$$
\mathcal{R}_\infty(\theta\mid q) = \frac{
\mathrm{E}_{o\sim \pi_\theta(\cdot\mid q)}[r(o\mid q)] - \mathrm{E}_{o\sim \pi_{\theta_{\old}}(o\mid q)}[r(o\mid q)]
}
{
\sigma(\pi_{\theta_{\old}}(\cdot\mid q))
}.
$$

\subsection{The reference-policy divergence penalty}

We consider the reference-policy divergence penalty in the GRPO's objective function given in Equation~(\ref{equ:Jgrpo}). To this end, for an arbitrary context $q$, we consider:
\begin{equation}
\mathcal{D}(\theta\mid q) \stackrel{\mathrm{def}}{=}  \mathbb{E}_{\{o_i\}_{i=1}^G \stackrel{\text{i.i.d.}}{\sim} \pi_{\theta_{\old}}(\cdot\mid q)}\left[\frac{1}{G}\sum_{i=1}^G D_i(\theta)\right].
\label{equ:dq}
\end{equation}
According to \cite{shao2024deepseekmathpushinglimitsmathematical}, the reference-policy divergence penalty is defined as an estimator of the KL divergence between $\pi_\theta(\cdot\mid q)$ and $\pi_\rf(\cdot\mid q)$, specifically using as inspiration an estimator discussed in \cite{schulman2020}. It can be readily observed that
$$
\mathcal{D}(\theta\mid q) = \mathrm{KL}_{0}(\pi_\theta(\cdot \mid q)\mid \mid \pi_{\rf}(\cdot\mid q);\pi_{\theta_{\old}}(\cdot\mid q))
$$
where 
$$
\mathrm{KL}_{0}(\pi \mid \mid \pi^*; \pi') \stackrel{\mathrm{def}}{=} \mathbb{E}_{x\sim \pi'}\left[\frac{\pi^*(x)}{\pi(x)}\right] - \mathbb{E}_{x\sim \pi'}\left[\log\frac{\pi^*(x)}{\pi(x)}\right] - 1.
$$

Indeed, the GRPO's reference-policy divergence penalty is an unbiased estimator of the KL divergence $\mathrm{KL}(\pi_\theta(\cdot\mid q)\mid\mid \pi_{\rf}(\cdot\mid q))$ in the case where $\pi_{\theta}(\cdot\mid q)=\pi_{\theta_{\old}}(\cdot\mid q)$, but not in general. More importantly, for optimisation purposes, it is the gradient of the reference policy divergence that matters, and the two divergences have different gradients. 

The gradient of $\mathrm{KL}_{0}(\pi_\theta(\cdot \mid q)\mid \mid \pi_{\rf}(\cdot\mid q);\pi_{\theta_{\old}}(\cdot\mid q))$ with respect to $\pi_\theta(\cdot\mid q)$ is given as:
\begin{equation}
\frac{\partial}{\partial \pi_{\theta}(o\mid q)} \mathrm{KL}_{0}(\pi_\theta(\cdot \mid q)\mid \mid \pi_{\rf}(\cdot\mid q);\pi_{\theta_{\old}}(\cdot\mid q)) = -\pi_{\theta_{\old}}(o \mid q)\frac{\pi_{\rf}(o\mid q)}{\pi_\theta(o\mid q)^2} + \pi_{\theta_{\old}}(o\mid q)\frac{1}{\pi_\theta(o\mid q)}.
\label{equ:gradKL0}
\end{equation}

For the KL divergence, we have
\begin{equation}
\frac{\partial}{\partial \pi_{\theta}(o\mid q)} \mathrm{KL}(\pi_\theta(\cdot\mid q)\mid \mid \pi_\rf(\cdot\mid q)) = -\log\frac{\pi_{\rf}(o\mid q)}{\pi_\theta(o\mid q)} + 1.
\label{equ:KLdiv}
\end{equation}

We observe that the gradients In Equations~(\ref{equ:gradKL0}) and (\ref{equ:KLdiv}) are different even in the case where $\pi_{\theta_{\old}}(\cdot\mid q) = \pi_{\theta}(\cdot\mid q)$, in which case
\begin{equation}
\frac{\partial}{\partial \pi_{\theta}(o\mid q)}\mathrm{KL}_{0}(\pi_\theta(\cdot \mid q)\mid \mid \pi_{\rf}(\cdot\mid q);\pi_{\theta_{\old}}(\cdot\mid q)) = -\frac{\pi_{\rf}(o\mid q)}{\pi_\theta(o\mid q)} + 1
\label{equ:KL0grad}
\end{equation}
which is linear in the probability ratio $\pi_{\rf}(o \mid q)/\pi_\theta(o\mid q)$, rather than logarithmic, as in the gradient of the KL divergence in Equation~(\ref{equ:KLdiv}). 

It is noteworthy that the gradient of the reference-policy divergence penalty, when $\pi_{\theta_\old}(\cdot\mid q) = \pi_{\theta}(\cdot\mid q)$, is equivalent to the gradient of the \emph{reverse KL divergence} between $\pi_\theta(\cdot\mid q)$ and $\pi_\rf(\cdot\mid q)$, i.e.,
$$
\mathrm{KL}_{\mathrm{Rev}}(\pi_\theta(\cdot\mid q)\mid \mid \pi_{\rf}(\cdot\mid q)) = \mathrm{KL}(\pi_{\rf}(\cdot\mid q)\mid\mid \pi_\theta(\cdot\mid q)) =\mathbb{E}_{o\sim \pi_\rf(\cdot\mid q)}\left[\log\frac{\pi_\rf(o\mid q)}{\pi_\theta(o\mid q)}\right],
$$
up to a non-essential additive constant. Indeed, it holds:
$$
\frac{\partial}{\partial \pi_\theta(o\mid q)}\mathrm{KL}_{\mathrm{rev}}(\pi_\theta(\cdot\mid q)\mid\mid \pi_\rf(\cdot\mid q)) = -\frac{\pi_\rf(o\mid q)}{\pi_\theta(o\mid q)}
$$
which is equal to the gradient in Equation~(\ref{equ:KL0grad}) up to an additive constant of value $1$. This additive constant is non-essential for determining stationary policies. 


\subsection{The alignment objective and stationary policies}

Having discussed the reward preference model and the reference-policy divergence penalty components of the GRPO's objective function, we now consider the objective function and its stationary policies. From our preceding discussion, we have:
$$
\mathcal{J}_{GRPO}(\theta) = \mathbb{E}_{q\sim \mu}[\mathcal{J}_{GRPO}(\pi_\theta(\cdot\mid q)\mid q)], 
$$
where
$$
\mathcal{J}_{GRPO}(\pi_\theta(\cdot\mid q)\mid q) = \mathbb{E}_{o\sim \pi_\theta(\cdot\mid q)}[\mathcal{P}_G(o\mid \pi_{\theta_{\old}}(\cdot\mid q),q)] - \beta\ \mathrm{KL}_{0}(\pi_\theta(\cdot\mid q)\mid \mid \pi_{\rf}(\cdot\mid q); \pi_{\theta_{\old}}(\cdot\mid q)).
$$

For each context $q$ and any previous policy $\pi_{\theta_\old}(\cdot\mid q)$, we consider the maxima of the following nonlinear programming problem:
$$
\begin{array}{rl}
\hbox{maximise} & \mathcal{J}_{GRPO}(\pi_\theta(\cdot\mid q)\mid q)\\
\hbox{over} & \pi_{\theta}(\cdot\mid q)\\
\hbox{subject to} & \pi_{\theta}(o\mid q)\geq 0, \forall o\\
& \sum_{o}\pi_{\theta}(o\mid q) = 1.
\end{array}
$$

Since our focus is on characterising stationary policies, we consider the maxima of the optimisation problem when $\pi_{\theta_\old}(\cdot\mid q) = \pi_\theta(\cdot\mid q)$. By the Karush-Kuhn-Tucker (KKT) optimality conditions~\citep{optbook}, for every output $o$ and context $q$, for each maximum, it either holds that $\pi_\theta(o\mid q) = 0$ or
\begin{equation}
\left(1-\frac{\mathcal{P}_G(o\mid \pi_{\theta}(\cdot\mid q),q) - \mathbb{E}_{o'\sim \pi_{\theta}(\cdot\mid q)}[\mathcal{P}_G(o'\mid \pi_{\theta}(\cdot\mid q),q)]}{\beta}\right)\pi_\theta(o\mid q) = \pi_{\rf}(o\mid q).
\label{equ:optcondition}
\end{equation}
The details are provided in Appendix~\ref{sec:kkt}. 

Note that for every context $q$ and output $o$ such that $\pi_\theta(o\mid q) > 0$, it must hold:
$$
\mathcal{P}_G(o\mid \pi_{\theta}(\cdot\mid q),q) < \mathbb{E}_{o'\sim \pi_{\theta}(\cdot\mid q)}[\mathcal{P}_G(o'\mid \pi_{\theta}(\cdot\mid q),q)] + \beta.
$$
Thus, the expected group-relative preference for any output selected with positive probability under a stationary policy is within an additive constant of $\beta$ of the expected group-relative preference of a randomly chosen output under the same stationary policy.

We can rewrite Equation~(\ref{equ:optcondition}) as:
$$
\pi_\theta(o\mid q) = g\left(\frac{\mathcal{P}_G(o\mid \pi_{\theta}(\cdot\mid q),q) - \mathbb{E}_{o'\sim \pi_{\theta}(\cdot\mid q)}[\mathcal{P}_G(o'\mid \pi_{\theta}(\cdot\mid q),q)]}{\beta}\right)\pi_\rf(o\mid q).
$$
where $g(x) \stackrel{\mathrm{def}}{=} 1/(1-x)$. We observe that the aggregation of preferences is different than logarithmic pooling. 

\paragraph{Groups of size two} For groups of size two, (\ref{equ:optcondition}) corresponds to: 
\begin{equation}
\left(1-\frac{\mathbb{P}_{o'\sim \pi_\theta(\cdot\mid q)}[r > r'\mid o]-\mathbb{P}_{o'\sim \pi_\theta(\cdot\mid q)}[r < r'\mid o]}{\beta}\right)\pi_\theta(o\mid q) = \pi_{\rf}(o\mid q)
\label{equ:optcondition2}
\end{equation}
where $r$ and $r'$ are respective rewards of outputs $o$ and $o'$, under context $q$.

\paragraph{The limit of large groups} For the limit of large group sizes, (\ref{equ:optcondition}) corresponds to:
\begin{equation}
\left(1-\frac{r(o\mid q) - \mathbb{E}_{o'\sim \pi_{\theta}(\cdot\mid q)}[r(o'\mid q)]}{\beta\sigma(\pi_\theta(\cdot\mid q))}\right)\pi_\theta(o\mid q) = \pi_{\rf}(o\mid q).
\label{equ:optconditioninf}
\end{equation}
Note that the scale normalisation of the rewards with the standard deviation $\sigma(\pi_\theta(\cdot\mid q))$ can be interpreted as using an effective regularisation constant of $\beta \sigma(\pi_\theta(\cdot\mid q))$ for the reference-penalty divergence penalty. For a policy $\pi_\theta(\cdot\mid q)$ that is more concentrated on placing its mass on a single output, the smaller the deviation $\sigma(\pi_\theta(\cdot\mid q))$, and, thus, a smaller effective weight is placed on the reference-policy divergence penalty than on the reward preference maximisation.   

Equation~(\ref{equ:optcondition}) is a fixed-point equation for $\pi_\theta(\cdot\mid q)$ for any group size, while (\ref{equ:optcondition2}) and (\ref{equ:optconditioninf}) are the corresponding conditions for the case of groups of size two and the limit of large groups. A distribution $\pi_\theta(\cdot\mid q)$ satisfying these fixed-point equations can be obtained in a closed-form in some cases. We demonstrate this in the next section for the case of binary questions, with groups of either size two or asymptotically large group size. This provides insights into some of the properties of the preference aggregation according to the GRPO criteria.  

\begin{figure}[t!]
\centering
\includegraphics[width=0.65\linewidth]{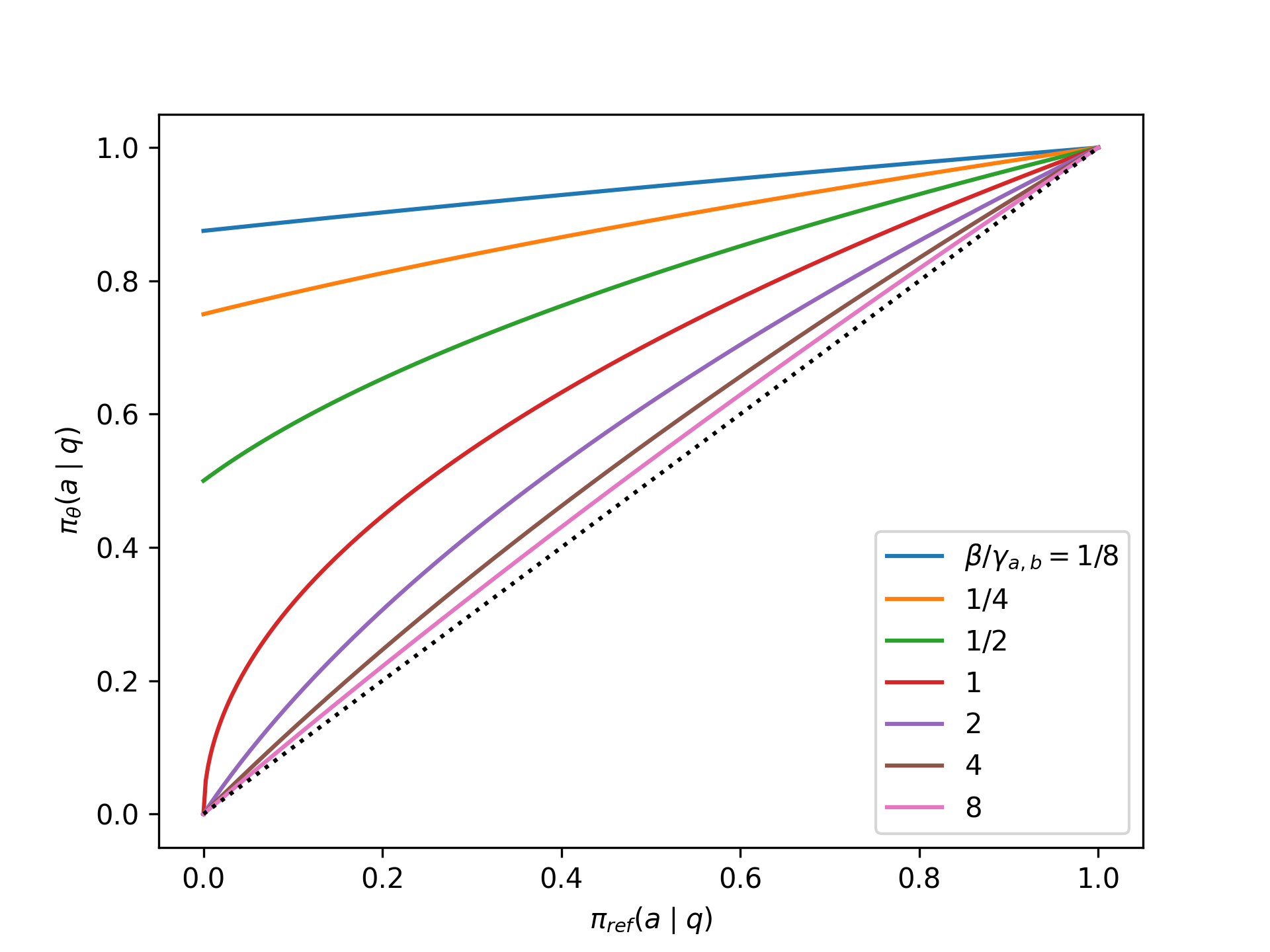}
\caption{GRPO's preference aggregation for the case of binary questions 
 with two answers, $a$ or $b$, and groups of size two: $\pi_\theta(a\mid q)$ versus $\pi_{\rf}(a\mid q)$ for the answer $a$ where $\mathcal{P}(a\succ b) > \mathcal{P}(b\succ a)$.}
\label{fig:bin}
\end{figure}

\subsubsection{Binary questions}
\label{sec:binary}

\paragraph{Groups of size two} Consider a question (context) $q$ that has two possible answers (outputs), $a$ or $b$, and groups of size two. Without loss of generality, assume that $\mathcal{P}(a\succ b\mid q) >  \mathcal{P}(b\succ a\mid q)$. Then, we have
\begin{equation}
\pi_\theta(a\mid q) = \frac{1}{2}\left(1-\frac{\beta}{\gamma_{a,b}}+\sqrt{\left(1-\frac{\beta}{\gamma_{a,b}}\right)^2 + 4\frac{\beta}{\gamma_{a,b}} \pi_{\rf}(a\mid q)}\right)
\label{equ:binary}
\end{equation}
where $\gamma_{a,b}\stackrel{\mathrm{def}}{=}\mathcal{P}(a\succ b\mid q)-\mathcal{P}(b\succ a\mid q)$ is the (signed) confidence margin of question answers.\footnote{For simplicity of notation, in $\gamma_{a,b}$, we omit the indication of dependency on the context $q$.} In the case of a tie, i.e., when $\mathcal{P}(a\succ b\mid q) =  \mathcal{P}(b\succ a\mid q)$, it holds $\pi_\theta(\cdot\mid q) =\pi_\rf(\cdot\mid q)$.

The value of $\pi_\theta(a\mid q)$ depends only on the ratio $\beta/\gamma_{a,b}$ and $\pi_{\rf}(a\mid q)$. We may regard the ratio $\beta/\gamma_{a,b}$ as the effective regularisation constant of the reference-policy divergence penalty. As expected, $\pi_\theta(a\mid q)$ is decreasing in $\beta$. It converges to the value $1$ as $\beta\rightarrow 0$, and converges to $\pi_{\rf}(a\mid q)$ as $\beta\rightarrow \infty$. Specifically, $\pi_\theta(a\mid q) = \sqrt{\pi_{\rf}(a\mid q)}$ for $\beta = \gamma_{a,b}$. Moreover, as expected, $\pi_\theta(a\mid q)$ increases in the confidence margin $\gamma_{a,b}$, as larger confidence margin means larger reward preference for answer $a$. As expected, the value of $\pi_\theta(a\mid q)$ increases in $\pi_\rf(a\mid q)$. It is noteworthy that this dependence is continuous except at $\pi_\rf(a\mid q) = 0$ where it is discontinuous whenever $\beta < \gamma_{a,b}$. Recall that when $\pi_\rf(a\mid q)=0$, then $\pi_\theta(a\mid q) = 0$ as the domain of $\pi_\theta(\cdot\mid q)$ is contained in the support of $\pi_\rf(\cdot\mid q)$. See Figure~\ref{fig:bin} for an illustration. 

It is worth noting that it holds 
$$
\pi_\theta(a\mid q) \geq  \max\left\{1 - \frac{\beta}{\gamma_{a,b}},\pi_\rf(a\mid q)\right\}.
$$
Hence, if $\beta$ is small enough relative to the confidence margin $\gamma_{a,b}$, the value of $\pi_\theta(a\mid q)$ is close to $1$, no matter what the value of $\pi_\rf(a\mid q)$ is. 

In the case of deterministic rewards such that $r(a\mid q) > r(b\mid q)$, $\mathcal{P}(a\succ b\mid q) = 1$, and thus $\gamma_{a,b} = 1$. The preference aggregation depends solely on the comparison of the rewards. This means that the relative preference between the two possible answers, $a$ and $b$, is determined by which one has the higher reward, rather than the absolute values of those rewards. Consequently, the model is invariant to the absolute magnitude of the rewards and is instead focused on the ranking of the rewards. This property is in line with the idea that the preference aggregation is driven by the order of rewards (i.e., a relative comparison), rather than their actual values. Thus, the model focuses on the relative ranking between alternatives and ignores any scale or offset in the reward values themselves.

\begin{figure}[t!]
\centering
\includegraphics[width=0.65\linewidth]{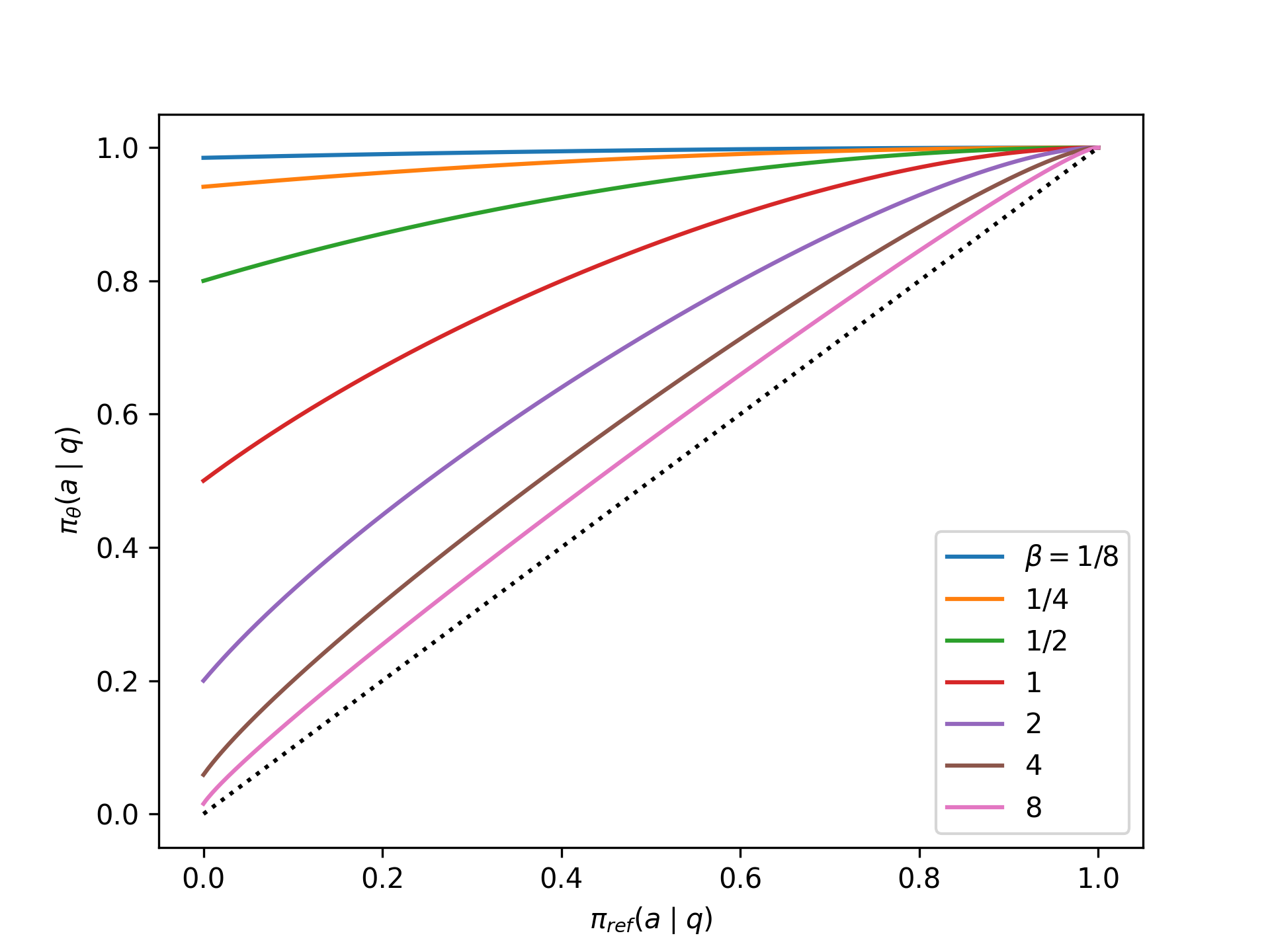}
\caption{GRPO's preference aggregation for the case of binary questions 
 with two answers, $a$ or $b$, in the limit of large group size: $\pi_\theta(a\mid q)$ versus $\pi_{\rf}(a\mid q)$ for the answer $a$ where $r(a\mid q) > r(b\mid q)$.}
\label{fig:bininf}
\end{figure}

\paragraph{The limit of large group size} For a question $q$ with answers $a$ or $b$ such that $r(a\mid q)>r(b\mid q)$, in the limit of large group size, we have
\begin{equation}
\pi_\theta(a\mid q) = \frac{2\beta^2 \pi_\rf(a\mid q)+1 + \sqrt{1+4\beta^2 \pi_\rf(a\mid q)(1-\pi_\rf(a\mid q))}}{2(1+\beta^2)}.
\label{equ:binaryinf}
\end{equation}
Intuitively, when $r(a\mid q) = r(b\mid q)$, then $\pi_\theta(\cdot\mid q) = \pi_\rf(\cdot\mid q)$.

A notable difference from the case of groups of size two is that the aggregation of preferences depends solely on $\pi_\rf(\cdot\mid q)$, the regularisation constant $\beta$, and the comparison of the expected rewards $r(a\mid q)$ and $r(b\mid q)$. The aggregation of preferences is more biased towards the reward preference than for the case of groups of size two. The dependence on $\pi_\rf(a\mid q)$ is discontinuous at $\pi_\rf(a\mid q)=0$ for every $\beta > 0$. See an illustration in Figure~\ref{fig:bininf}. It can be readily noted that
$$
\pi_\theta(a\mid q) \geq \max\left\{\frac{1}{1+\beta^2}, \pi_\rf(a\mid q)\right\}.
$$
Comparing with the bound for groups of size two, for the case of deterministic rewards, we observe that $\pi_\theta(q\mid q)$ is now lower bounded by $1/(1+\beta^2)$ while for the case of groups of size two, it is lower bounded by $1-\beta$. The lower bound $1/(1+\beta^2)$ is larger than $1-\beta$, for every $\beta > 0$.

\section{Extensions}

The GRPO's alignment objective can naturally be extended in different directions by redefining the reward preference model or the reference-policy divergence penalty. Here, we discuss some different variants. 

\paragraph{Using the direct KL divergence penalty} As noted, as far as the stationary policies are concerned, the GRPO's reference-policy divergence penalty is essentially the reverse KL divergence between $\pi_{\theta}(\cdot\mid q)$ and $\pi_\rf(\cdot\mid q)$. We can easily convert the reference-policy divergence penalty to correspond to the direct KL divergence between the two distributions. This can be done by the standard importance sampling trick for Monte Carlo estimation by redefining the penalty terms in the GRPO objective as follows:
$$
D_i(\theta) = \frac{\pi_\theta(o_i\mid q)}{\pi_{\theta_{\old}}(o_i\mid q)}\left(\frac{\pi_{\rf}(o_i\mid q)}{\pi_\theta(o_i\mid q)} - \log \frac{\pi_{\rf}(o_i\mid q)}{\pi_\theta(o_i\mid q)} -1\right).
$$
With this new definition, the expected reference-policy divergence penalty defined in Equation~(\ref{equ:dq}) corresponds to the KL divergence between $\pi_\theta(\cdot\mid q)$ and $\pi_\rf(\cdot\mid q)$, i.e. $\mathcal{D}(\theta\mid q) = \mathrm{KL}(\pi_\theta(\cdot\mid q)\mid\mid \pi_{\rf}(\cdot\mid q))$. 

The resulting aggregation of preferences satisfies:
$$
\pi_\theta(o\mid q) = \frac{1}{Z_q} e^{\frac{\mathcal{P}_G(o\mid \pi_\theta(\cdot\mid q),q)}{\beta}}\pi_\rf(o\mid q)
$$
where $Z_q$ is a normalisation constant. 

\begin{figure}[t!]
\centering
\includegraphics[width=0.65\linewidth]{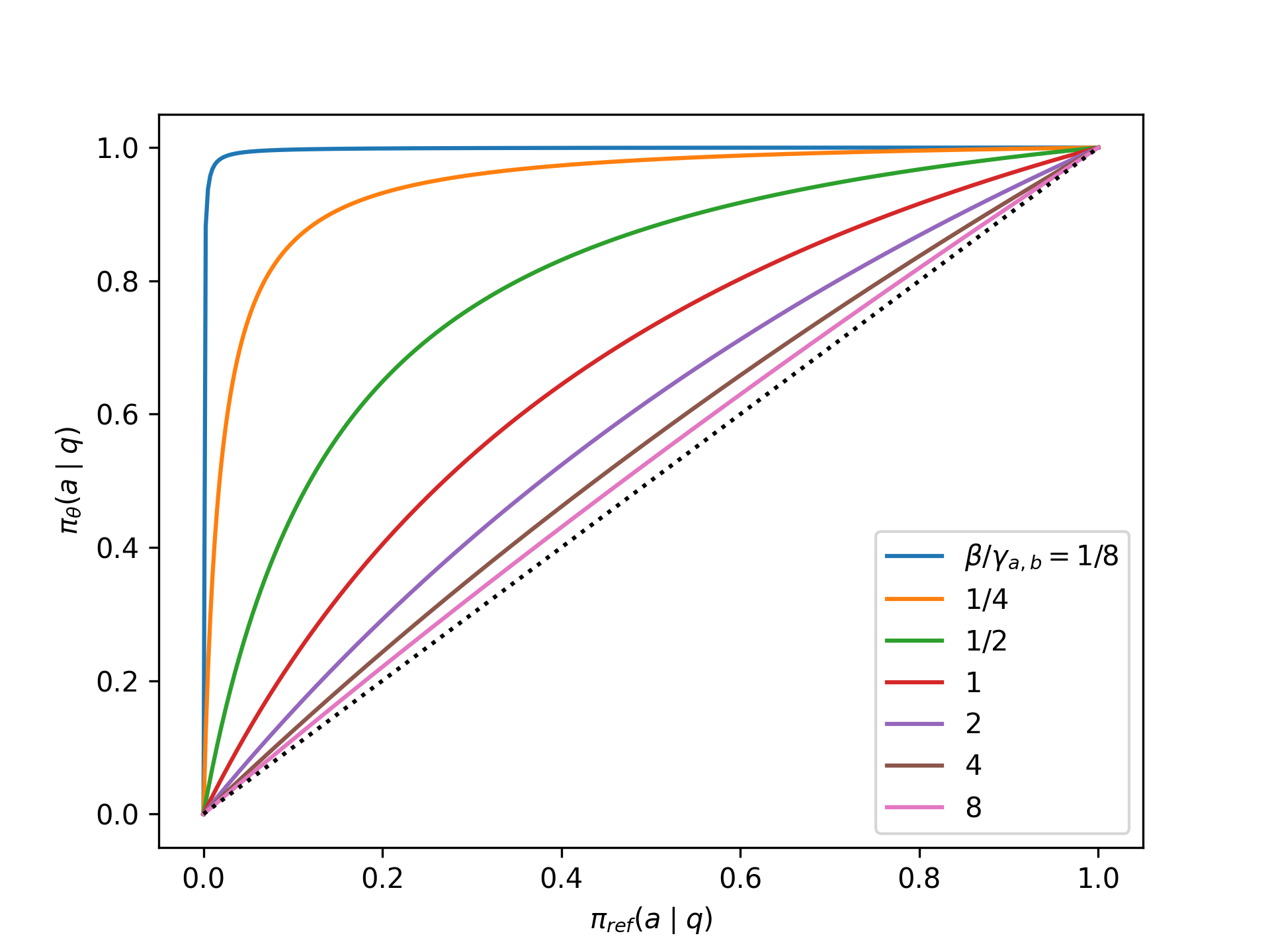}
\caption{Preference aggregation according to GRPO's reward preference model and direct KL divergence penalty, for the case of binary questions 
 with two possible answers, $a$ or $b$, and groups of size two: $\pi_\theta(a\mid q)$ versus $\pi_{\rf}(a\mid q)$ for the answer $a$ where $r(a\mid q) > r(b\mid q)$. A notable difference from the GRPO's alignment results, shown in Figure~\ref{fig:bin}, is a lack of discontinuity at $\pi_\rf(a\mid q) = 0$.}
\label{fig:bin2KLdirect}
\end{figure}

For groups of size two, this aggregation of preference is akin to the NLHF alignment objective. Specifically, for binary questions, with two possible answers, $a$ or $b$, it holds
$$
\pi_\theta(a\mid q) = \frac{1}{Z_q} e^{\frac{\gamma_{a,b}}{2\beta}}\pi_\rf(a\mid q)
$$
where $Z_q$ is a normalisation constant, given as $Z_q = e^{\gamma_{a,b}/(2\beta)} \pi_\rf(a\mid q) + e^{-\gamma_{a,b}/(2\beta)}(1-\pi_\rf(a\mid q))$. Here, recall that $\gamma_{a,b} = \mathcal{P}(a\succ b\mid q) - \mathcal{P}(b\succ a\mid q)$, which we first defined in Section~\ref{sec:binary}. See Figure~\ref{fig:bin2KLdirect} for an illustration.

For the limit of a large group size, given $a$ and $b$ such that $r(a\mid q)>r(b\mid q)$, the optimal value of $\pi_\theta(a\mid q)$ is $1$ if $\beta$ is sufficiently small. If $\beta$ is sufficiently large, there exist two values of $\pi_\theta(a\mid q)$ that maximise the objective function if $\beta$ is sufficiently large. See Appendix~\ref{sec:binaryKL} for details.

\paragraph{Shift-only normalisation} Consider the GRPO's reward preference model with shift-normalised rewards and without using scale-normalisation. Hence, we consider the advantage terms defined as $A_i = r_i - \mathrm{mean}(r_1, \ldots, r_G)$. Then, we have
$$
\mathbb{E}_{o\sim \pi_\theta(\cdot\mid q)}[\mathcal{P}_G(o\mid \pi_{\theta_{\old}}(\cdot\mid q),q)]  = \left(1-\frac{1}{G}\right)(\mathbb{E}_{o\sim \pi_\theta(\cdot\mid q)}[r(o\mid q)]- \mathbb{E}_{o\sim \pi_{\theta_\old}(\cdot\mid q)}[r(o\mid q)]).
$$
This results in a reward preference model similar to that of the RLHF alignment approach, but replacing the reward model with a sample mean estimate of the rewards. Combining the two extensions, we obtain aggregation of preferences similar to that of the RLHF alignment objective. 

\bibliographystyle{abbrvnat}
\bibliography{ref}

\appendix

\section{Stationary policies}
\label{sec:kkt}

We consider the following optimisation problem:
$$
\begin{array}{rl}
\hbox{maximise} & \mathcal{J}_{GRPO}(\pi_\theta(\cdot\mid q)\mid q)\\
\hbox{over} & \pi_{\theta}(\cdot\mid q)\\
\hbox{subject to} & \pi_{\theta}(o\mid q)\geq 0, \forall o\\
& \sum_{o}\pi_{\theta}(o\mid q) = 1.
\end{array}
$$

By the KKT conditions, if $\pi_\theta(\cdot\mid q)$ is a local optimum, then there exist constants $\gamma_o$ and $\lambda$ such that, for all $o$, the following conditions hold:
$$
\begin{array}{rl}
\hbox{stationarity:} & \frac{\partial}{\partial \pi_{\theta}(o\mid q)}\mathcal{J}_{GRPO}(\pi_\theta(\cdot\mid q)\mid q) - \lambda + \gamma_o = 0,\\ 
\hbox{non-negativity:} & \gamma_0\geq 0, \hbox{and},\\
\hbox{complementary slackness:} & \gamma_o \pi_{\theta}(o\mid q) = 0.
\end{array}
$$

Therefore, either $\pi_\theta(o\mid q) = 0$, or $\pi_\theta(o\mid q) > 0$ and
$$
\mathcal{P}_G(o\mid \pi_{\theta_{\old}}(\cdot\mid q),q) -\beta \left(-\pi_{\theta_{\old}}(o \mid q)\frac{\pi_{\rf}(o\mid q)}{\pi_\theta(o\mid q)^2} + \pi_{\theta_{\old}}(o\mid q)\frac{1}{\pi_\theta(o\mid q)}\right) -\lambda = 0.
$$
Under the condition $\pi_\theta(\cdot \mid q) = \pi_{\theta_{\old}}(\cdot\mid q)$, we have
$$
\pi_{\rf}(o\mid q) = \left(1-\frac{\mathcal{P}_G(o\mid \pi_{\theta}(\cdot\mid q),q)-\lambda}{\beta}\right)\pi_\theta(o\mid q).
$$

By using the constraint $\sum_{o'}\pi_\theta(o'\mid q) = 1$, we have
$$
\left(1-\frac{\mathcal{P}_G(o\mid \pi_{\theta}(\cdot\mid q),q) - \mathbb{E}_{o'\sim \pi_{\theta}(\cdot\mid q)}[\mathcal{P}_G(o'\mid \pi_{\theta}(\cdot\mid q),q)]}{\beta}\right)\pi_\theta(o\mid q) = \pi_{\rf}(o\mid q),
$$
which is the condition shown in Equation~(\ref{equ:optcondition}).

\section{Binary questions}

\subsection{The GRPO's alignment objective}

\paragraph{Groups of size two} We first consider the reward preference model part of the objective. Note that
\begin{eqnarray*}
\mathbb{E}_{o\sim \pi_\theta(\cdot\mid q)}[\mathcal{P}_2(o\mid \pi_{\theta_{\old}}(\cdot\mid q),q)] & = & \pi_\theta(a\mid q)\pi_{\theta_\old}(b\mid q)\mathbb{E}[\mathrm{sign}(r_1-r_2)\mid o_1 = a, o_2 = b, q]\\
&& + \pi_\theta(b\mid q)\pi_{\theta_\old}(a\mid q)\mathbb{E}[\mathrm{sign}(r_1-r_2)\mid o_1 = b, o_2 = a, q]\\
& = & \pi_\theta(a\mid q)\pi_{\theta_\old}(b\mid q)\mathbb{E}[\mathrm{sign}(r_1-r_2)\mid o_1 = a, o_2 = b, q]\\
&& - \pi_\theta(b\mid q)\pi_{\theta_\old}(a\mid q)\mathbb{E}[\mathrm{sign}(r_2-r_1)\mid o_1 = b, o_2 = a, q]\\
& = & \pi_\theta(a\mid q)\pi_{\theta_\old}(b\mid q)\mathbb{E}[\mathrm{sign}(r_1-r_2)\mid o_1 = a, o_2 = b, q]\\
&& - \pi_\theta(b\mid q)\pi_{\theta_\old}(a\mid q)\mathbb{E}[\mathrm{sign}(r_1-r_2)\mid o_1 = a, o_2 = b, q]\\
&=& (\pi_\theta(a\mid q)\pi_{\theta_\old}(b\mid q)-\pi_\theta(b\mid q)\pi_{\theta_\old}(a\mid q))\times \\
&& \times \mathbb{E}[\mathrm{sign}(r_1-r_2)\mid o_1 = a, o_2 = b, q].
\end{eqnarray*}

Next, note that
\begin{eqnarray*}
\pi_\theta(a\mid q)\pi_{\theta_\old}(b\mid q)-\pi_\theta(b\mid q)\pi_{\theta_\old}(a\mid q) &=& \pi_\theta(a\mid q)(1-\pi_{\theta_\old}(a\mid q))-(1-\pi_\theta(a\mid q))\pi_{\theta_\old}(a\mid q)\\
&=& \pi_\theta(a\mid q) - \pi_{\theta_\old}(a\mid q),
\end{eqnarray*}
and
$$
\mathbb{E}[\mathrm{sign}(r_1-r_2)\mid o_1 = a, o_2 = b, q] 
= \gamma_{a,b},
$$
where 
$$
\gamma_{a,b}\stackrel{\mathrm{def}}{=}\mathcal{P}(a\succ b\mid q) - \mathcal{P}(b\succ a\mid q).
$$

Hence, we have
\begin{equation}
\mathbb{E}_{o\sim \pi_\theta(\cdot\mid q)}[\mathcal{P}_2(o\mid \pi_{\theta_{\old}}(\cdot\mid q),q)] = \gamma_{a,b}(\pi_\theta(a\mid q) - \pi_{\theta_\old}(a\mid q)).
\label{equ:binaryreward2}
\end{equation}

Combining this with the reference-policy divergence penalty, we have
\begin{eqnarray*}
\mathcal{J}_{GRPO}(\pi_\theta(a\mid q)\mid q)& =& \gamma_{a,b}\pi_\theta(a\mid q)\\
&& - \beta\left(\pi_{\theta_\old}(a\mid q)\frac{\pi_\rf(a\mid q)}{\pi_\theta(a\mid q)} + (1-\pi_{\theta_\old}(a\mid q))\frac{1-\pi_\rf(a\mid q)}{1-\pi_\theta(a\mid q)}
\right .\\
&&\left . + \pi_{\theta_\old}(a\mid q)\log(\pi_\theta(a\mid q)) + (1-\pi_{\theta_\old}(a\mid q))\log(1-\pi_\theta(a\mid q))\right) + \hbox{ const}.
\end{eqnarray*}

Taking the first derivative with respect to $\pi_\theta(a\mid q)$, we obtain:
\begin{eqnarray*}
\frac{d\mathcal{J}_{GRPO}(\pi_\theta(a\mid q)\mid q)}{d\pi_\theta(a\mid q)} & =& \gamma_{a,b}\\
&& - \beta\left(-\frac{\pi_{\theta_{\old}}(a\mid q)\pi_{\rf}(a\mid q)} {\pi_\theta(a\mid q)^2} + \frac{(1-\pi_{\theta_{\old}}(a\mid q))(1-\pi_{\rf}(a\mid q))}{(1-\pi_\theta(a\mid q))^2} \right .\\
&& \left . + \frac{\pi_{\theta_{\old}}(a\mid q)}{\pi_\theta(a\mid q)} - \frac{1-\pi_{\theta_{\old}}(a\mid q)}{1-\pi_\theta(a\mid q)} \right),
\end{eqnarray*}
which for the case where $\pi_{\theta_{\old}}(\cdot \mid q) = \pi_\theta(\cdot\mid q)$ simplifies to:
$$
\frac{d}{d\pi_\theta(a\mid q)}\mathcal{J}_{GRPO}(\pi_\theta(a\mid q)\mid q) = \gamma_{a,b} - \beta \frac{\pi_\theta(a\mid q)-\pi_{\rf}(a\mid q)}{\pi_\theta(a\mid q)(1-\pi_\theta(a\mid q))}.
$$
Setting the derivative to zero, we obtain
\begin{equation}
\beta \frac{\pi_\theta(a\mid q)-\pi_{\rf}(a\mid q)}{\pi_\theta(a\mid q)(1-\pi_\theta(a\mid q))} = \gamma_{a,b}.
\label{equ:cond0}
\end{equation}
Now, clearly for the case where $\gamma_{a,b} = 0$, we have $\pi_\theta(a\mid q) = \pi_\rf(a\mid q)$. For the case where $\gamma_{a,b}\neq 0$, by simple rearrangements, it can be shown that Equation~(\ref{equ:cond0}) is equivalent to the following quadratic equation:
$$
\pi_\theta(a\mid q)^2 - \left(1-\frac{\beta}{\gamma_{a,b}}\right)\pi_\theta(a\mid q) - \frac{\beta}{\gamma_{a,b}}\pi_\rf(a\mid q) = 0.
$$
For the case $\gamma_{a,b} >0$, this quadratic equation has a unique non-negative solution given as:
$$
\pi_\theta(a\mid q) = \frac{1}{2}\left(\sqrt{\left(1-\frac{\beta}{\gamma_{a,b}}\right)^2 + 4\frac{\beta}{\gamma_{a,b}} \pi_{\rf}(a\mid q)}+\left(1-\frac{\beta}{\gamma_{a,b}}\right)\right)
$$
which corresponds to the asserted equation in Equation~(\ref{equ:binary}).

\paragraph{The limit of large group size} Note that

$$
r(a\mid q) - \mathbb{E}_{o\sim \pi_{\theta_{\old}}(\cdot\mid q)}[r(o\mid q)] = (1-\pi_{\theta_{\old}}(a\mid q))(r(a\mid q)-r(b\mid q)),
$$

$$
r(b\mid q) - \mathbb{E}_{o\sim \pi_{\theta_{\old}}(\cdot\mid q)}[r(o\mid q)] = \pi_{\theta_{\old}}(a\mid q)(r(b\mid q)-r(a\mid q)),
$$
and
$$
\sigma(\pi_{\theta_{\old}}(\cdot \mid q))^2 = (r(a\mid q)-r(b\mid q))^2 \pi_{\theta_{\old}}(a\mid q)(1-\pi_{\theta_{\old}}(a\mid q)).
$$

It follows that
$$
\frac{r(a\mid q) - \mathbb{E}_{o\sim \pi_{\theta_{\old}}(\cdot\mid q)}[r(o\mid q)]}{\sigma(\pi_{\theta_{\old}}(\cdot \mid q))} = \sqrt{\frac{1-\pi_{\theta_{\old}}(a\mid q)}{\pi_{\theta_{\old}}(a\mid q)}}\mathrm{sign}(r(a\mid q)-r(b\mid q)).
$$
and
$$
\frac{r(b\mid q) - \mathbb{E}_{o\sim \pi_{\theta_{\old}}(\cdot\mid q)}[r(o\mid q)]}{\sigma(\pi_{\theta_{\old}}(\cdot \mid q))} = -\sqrt{\frac{\pi_{\theta_{\old}}(a\mid q)}{1-\pi_{\theta_{\old}}(a\mid q)}}\mathrm{sign}(r(a\mid q)-r(b\mid q)).
$$
Hence, we have
\begin{eqnarray}
&& \frac{\mathbb{E}_{o\sim \pi_\theta(\cdot\mid q)}[r(o\mid q)] - \mathbb{E}_{o\sim \pi_{\theta_{\old}}(\cdot\mid q)}[r(o\mid q)]}{\sigma(\pi_{\theta_{\old}}(\cdot \mid q))}\nonumber\\
&=& \left(\frac{1}{\sqrt{\pi_{\theta_\old}(a\mid q)(1-\pi_{\theta_\old}(a\mid q))}}\pi_\theta(a\mid q) - \sqrt{\frac{\pi_{\theta_{\old}}(a\mid q)}{1-\pi_{\theta_{\old}}(a\mid q)}}\right)\mathrm{sign}(r(a\mid q)-r(b\mid q)).\label{equ:binaryinfreward}
\end{eqnarray}

The objective function is as follows:
\begin{eqnarray*}
\mathcal{J}_{GRPO}(\pi_\theta(a\mid q)\mid q) &=& \frac{1}{\sqrt{\pi_{\theta_\old}(a\mid q)(1-\pi_{\theta_\old}(a\mid q))}}\mathrm{sign}(r(a\mid q)-r(b\mid q))\pi_\theta(a\mid q)\\
&& - \beta\left(\frac{\pi_{\theta_{\old}}(a\mid q)\pi_{\rf}(a\mid q)} {\pi_\theta(a\mid q)} + \frac{(1-\pi_{\theta_{\old}}(a\mid q))(1-\pi_{\rf}(a\mid q))}{1-\pi_\theta(a\mid q)} \right .\\
&& \left . + \pi_{\theta_{\old}}(a\mid q)\log(\pi_\theta(a\mid q)) + (1-\pi_{\theta_{\old}}(a\mid q))\log(1-\pi_\theta(a\mid q))) \right) + \hbox{ const}.
\end{eqnarray*}

The derivative of $\mathcal{J}_{GRPO}(\pi_\theta(\cdot\mid q))$ with respect to $\pi_\theta(a\mid q)$, evaluated at $\pi_\theta(a\mid q)$ such that $\pi_{\theta_{\old}}(\cdot \mid q) = \pi_\theta(\cdot\mid q)$, is equal to:
\begin{eqnarray*}
\frac{d}{d\pi_\theta(a\mid q)}\mathcal{J}_{GRPO}(\pi_\theta(a\mid q)\mid q) & = &  \frac{1}{\sqrt{\pi_{\theta_\old}(a\mid q)(1-\pi_{\theta_\old}(a\mid q))}}\mathrm{sign}(r(a\mid q)-r(b\mid q)) \\
&& - \beta \frac{\pi_\theta(a\mid q)-\pi_{\rf}(a\mid q)}{\pi_\theta(a\mid q)(1-\pi_\theta(a\mid q))}
\end{eqnarray*}
which when set to zero yields
$$
\beta(\pi_\theta(a\mid q)-\pi_{\rf}(a\mid q)) = \sqrt{\pi_\theta(a\mid q)(1-\pi_\theta(a\mid q))}\mathrm{sign}(r(a\mid q)-r(b\mid q)).
$$
Clearly, if $r(a\mid q) = r(b\mid q)$, then $\pi_\theta(a\mid q) = \pi_\rf(a\mid q)$. If $r(a\mid q) > r(b\mid q)$, then \begin{equation}
\beta(\pi_\theta(a\mid q)-\pi_{\rf}(a\mid q)) = \sqrt{\pi_\theta(a\mid q)(1-\pi_\theta(a\mid q))}.
\label{equ:cond1}
\end{equation}
This is equivalent to the following quadratic equation:
$$
(1+\beta^2)\pi_\theta(a\mid q)^2 - (2\beta^2 \pi_\rf(a\mid q)+1)\pi_\theta(a\mid q) + \beta^2 \pi_\rf(a\mid q)^2 = 0.
$$
Since by (\ref{equ:cond1}), $\pi_\theta(a\mid q)\geq \pi_\rf(a\mid q)$, the quadratic equation has a unique solution satisfying the latter condition, which is given as follows:
$$
\pi_\theta(a\mid q) = \frac{2\beta^2 \pi_\rf(a\mid q)+1 + \sqrt{1+4\beta^2 \pi_\rf(a\mid q)(1-\pi_\rf(a\mid q))}}{2(1+\beta^2)}.
$$
This shows that Equation~(\ref{equ:binaryinf}) holds.

\subsection{Using direct KL divergence penalty}
\label{sec:binaryKL}

\paragraph{Groups of size two} We consider the GRPO reward preference model with the reference-policy divergence penalty according to the KL divergence between $\pi_\theta(\cdot\mid q)$ and $\pi_\rf(\cdot \mid q)$. The reward preference part of the objective is as given in Equation~(\ref{equ:binaryreward2}). The objective function is given as follows:
$$
\mathcal{J}(\pi_\theta(a\mid q)\mid q) = \gamma_{a,b}\pi_\theta(a\mid q) - \beta \mathrm{KL}(\pi_\theta(a\mid q)\mid\mid \pi_\rf(a\mid q))) + \hbox{ const}
$$
where $K(p\mid\mid p')$ denotes the KL divergence between two Bernoulli distributions with means $p$ and $p'$.

It readily follows that
$$
\frac{d}{d\pi_\theta(a\mid q)} \mathcal{J}(\pi_\theta(\cdot\mid q)\mid q) = \gamma_{a,b} - \beta\left(\log\left((\frac{\pi_\theta(a\mid q)}{\pi_\rf(a\mid q)}\right)-\log\left(\frac{1-\pi_\theta(a\mid q)}{1-\pi_\rf(a\mid q)}\right)\right).
$$
By setting the derivative to zero, we obtain
$$
\pi_\theta(a\mid q) = \frac{1}{Z_q} e^{\frac{\gamma_{a,b}}{2\beta}}\pi_\rf(a\mid q)
$$
where $Z_q$ is the normalisation constant, given as $Z_q = e^{\gamma_{a,b}/(2\beta)} \pi_\rf(a\mid q) + e^{-\gamma_{a,b}/(2\beta)}(1-\pi_\rf(a\mid q))$.

\paragraph{The limit of large group size} In this case, the reward preference model component of the objective is as given in Equation~(\ref{equ:binaryinfreward}). The objective function is given as:
$$
\mathcal{J}(\pi_\theta(a\mid q)\mid q) = \frac{\tilde{\gamma}_{a,b}}{\sqrt{\pi_{\theta_\old}(a\mid q)(1-\pi_{\theta_\old}(a\mid q))}}\pi_\theta(a\mid q) -\beta\mathrm{KL}(\pi_\theta(a\mid q)\mid\mid \pi_\rf(a\mid q)) + \hbox{const}
$$
where $\tilde{\gamma}_{a,b} = \mathrm{sign}(r(a\mid b)>r(b\mid q))$.

The derivative of $\mathcal{J}(\pi_\theta(a\mid q)\mid q)$ with respect to $\pi_\theta(a\mid q)$, is given as:
$$
\frac{d \mathcal{J}(\pi_\theta(a\mid q)\mid q)}{d\pi_\theta(a\mid q)} = \frac{\tilde{\gamma}_{a,b}}{\sqrt{\pi_{\theta_\old}(a\mid q)(1-\pi_{\theta_\old}(a\mid q))}} -\beta\left(\log\left(\frac{\pi_\theta(a\mid q)}{1-\pi_\theta(a\mid q)}\right)-\log\left(\frac{\pi_\rf(a\mid q)}{1-\pi_\rf(a\mid q)}\right)\right).
$$


Without loss of generality, consider the case where $\tilde{\gamma}_{a,b} = 1$. Under the condition $\pi_{\theta_\old}(\cdot\mid q) = \pi_{\theta}(\cdot\mid q)$, we obtain:
$$
\frac{d}{d\pi_\theta(a\mid q)}\mathcal{J}(\pi_\theta(a\mid q)\mid q) = \beta h(\pi_\theta(a\mid q))
$$
where
$$
h(x) = \frac{1}{\beta}\frac{1}{\sqrt{x(1-x)}} - \log\left(\frac{x}{1-x}\right) + \log\left(\frac{\pi_\rf(a\mid q)}{1-\pi_\rf(a\mid q)}\right).
$$
It can be readily verified that the function $h(x)$ decreases on $(0,x^*]$ and increases on $[x^*,1)$ where $x^* = (1+\sqrt{1-1/(1+\beta^2)})/2$. Moreover, $\lim_{x\uparrow 0}h(x) = \infty$ and $\lim_{x\downarrow 1} h(x) = \infty$. 

If $\beta$ is small enough, then $h(x)>0$ for every $x\in [0,1]$. In this case, the objective function is maximised at $\pi_\theta(a\mid q) = 1$. On the other hand, if $\beta$ is sufficiently large, then there exist two values of $\pi_\theta(a\mid q)$ that satisfy $h(\pi_\theta(a\mid q)) = 0$. 

\end{document}